\documentclass[10pt,twocolumn,letterpaper]{article}

 \usepackage{cvpr}              
\usepackage{amsmath, bm}
\usepackage{multirow}
\usepackage{diagbox}
\usepackage{booktabs}
\definecolor{cvprblue}{rgb}{0.21,0.49,0.74}
\usepackage[pagebackref,breaklinks,colorlinks,allcolors=cvprblue]{hyperref}

\def\paperID{39033} 
\def\confName{CVPR}
\def\confYear{2026}

\title{CTFS : Collaborative Teacher Framework for Forward-Looking Sonar Image Semantic Segmentation with Extremely Limited Labels}

\author{
Ping Guo\textsuperscript{1}, Chengzhou Li\textsuperscript{1}, Guanchen Meng\textsuperscript{1}, Qi Jia\textsuperscript{1}\thanks{\raggedright Corresponding authors. Emails: \texttt{jiaqi@dlut.edu.cn}, \texttt{xin.fan@dlut.edu.cn}}, Jinyuan Liu\textsuperscript{1}\\
Zhu Liu\textsuperscript{1}, Yu Liu\textsuperscript{1}, Zhongxuan Luo\textsuperscript{1}, Xin Fan\textsuperscript{1}\footnotemark[1]\\[0.3em]
\textsuperscript{1}School of Software Technology, Dalian University of Technology, Dalian, China
}

\begin{document}
\maketitle

\begin{abstract}
As one of the most important underwater sensing technologies, forward-looking sonar exhibits unique imaging characteristics. Sonar images are often affected by severe speckle noise, low texture contrast, acoustic shadows, and geometric distortions. These factors make it difficult for traditional teacher–student frameworks to achieve satisfactory performance in sonar semantic segmentation tasks under extremely limited labeled data conditions. To address this issue, we propose a Collaborative Teacher Semantic Segmentation Framework for forward-looking sonar images. This framework introduces a multi-teacher collaborative mechanism composed of one general teacher and multiple sonar-specific teachers. By adopting a multi-teacher alternating guidance strategy, the student model can learn general semantic representations while simultaneously capturing the unique characteristics of sonar images, thereby achieving more comprehensive and robust feature modeling. Considering the challenges of sonar images, which can lead teachers to generate a large number of noisy pseudo-labels, we further design a cross-teacher reliability assessment mechanism. This mechanism dynamically quantifies the reliability of pseudo-labels by evaluating the consistency and stability of predictions across multiple views and multiple teachers, thereby mitigating the negative impact caused by noisy pseudo-labels. Notably, on the FLSMD dataset, when only 2\% of the data is labeled, our method achieves a 5.08\% improvement in mIoU compared to other state-of-the-art approaches.
\end{abstract}

\begin{center}
\small
\textbf{Code:} \url{https://github.com/pingggg516/CTFS}
\end{center}

\begin{figure}[t]
    \centering
    \includegraphics[width=\linewidth]{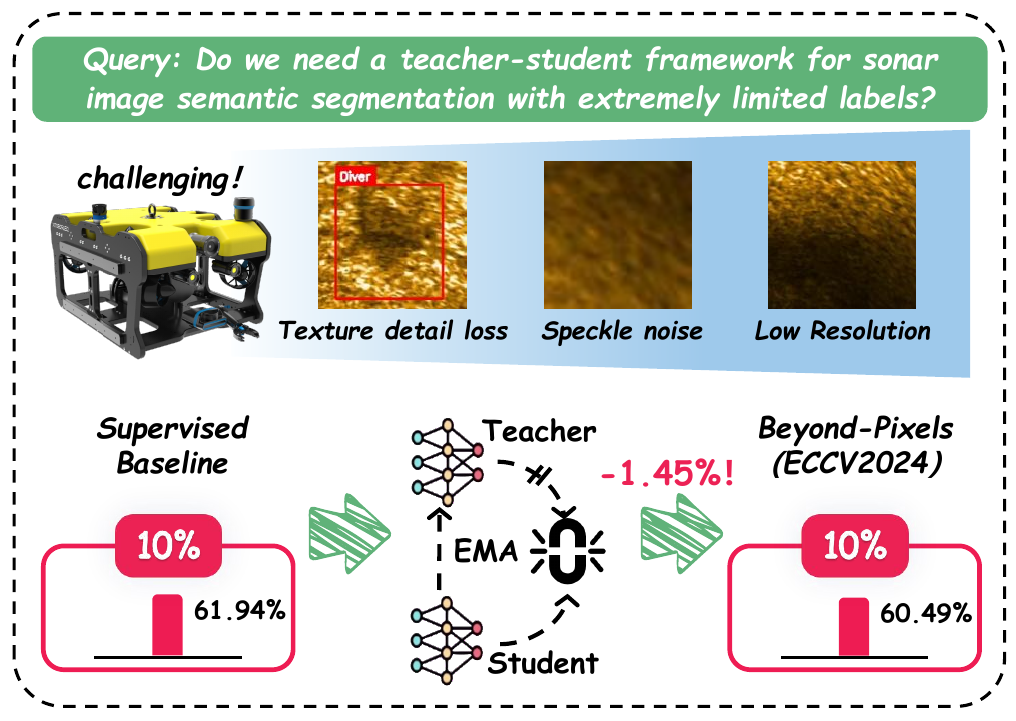}  
    \caption{The unique characteristics of sonar images lead to limited adaptability in conventional teacher–student frameworks, making them ineffective for handling semantic segmentation tasks under extremely scarce annotation conditions.}

\label{fig:Sonar1}
\end{figure}

\section{Introduction}
\label{sec:intro}

Seawater exhibits strong absorption of visible light, and in turbid waters, deep sea environments, or nighttime conditions, visible images become nearly ineffective~\cite{xie2022dataset, cao2022dynamic,li2025physics}. In contrast, sonar is inherently insensitive to variations in illumination and experiences minimal propagation loss in water, enabling it to capture targets over distances ranging from tens to hundreds of meters. Consequently, sonar has been widely applied in various underwater tasks, such as rescue operations, equipment inspection, and marine biological surveyss~\cite{liu2024two,mccann2018underwater,wang2025sonar}. Despite the numerous advantages of sonar technology, sonar images often suffer from issues such as blurred target boundaries, low resolution, and significant differences in target morphology compared to visible images~\cite{yuan2025improved,zhu2025saliency,li2024rgb}. Meanwhile, sonar data are typically highly specialized and confidential, making the manual annotation process both time-consuming and costly, which in turn limits the availability of high quality labeled samples for supervised learning methods.

To address the above issues, we introduce a teacher–student framework~\cite{cheng2025cgmatch,assefa2025dycon,Li_Guo_Meng_Jia_Liu_Liu_Liu_Liu_Luo_Fan_2026} based on semi-supervised learning to tackle the task of sonar semantic segmentation with extremely limited labels~\cite{xu2021end,na2023switching,chen2024virtual}. However, as shown in Figure~\ref{fig:Sonar1}, due to the inherent characteristics of sonar images—such as sparse texture details, severe speckle noise, and low resolution—the pseudo-label generation process of the teacher model is disturbed, resulting in a large number of low-quality pseudo-labels. Consequently, some algorithms even perform worse than the supervised baseline when using only 10\% of the labeled data. Moreover, existing methods generally lack teacher model designs that account for the unique characteristics of sonar imagery and thus fail to fully leverage the distinctive properties of sonar image~\cite{howlader2024beyond}. Such neglect undermines the effectiveness of the teacher–student framework and may even lead to a noticeable performance degradation compared with baseline methods when a large amount of unlabeled data is incorporated.

In this paper, we propose a Collaborative Teacher Framework for Sonar images (CTFS), designed to enhance semantic segmentation performance in scenarios where labeled samples are extremely scarce. First, we construct a multi-teacher collaborative learning mechanism that effectively transfers multi-perspective knowledge from different types of teacher models—including both general teacher and sonar-specific teachers to the student model, enabling it to better capture and utilize the unique characteristics of sonar imagery. Considering that the inherent characteristics of sonar images often lead to the generation of numerous noisy pseudo-labels, we further design a multi-view pseudo-label evaluation algorithm to dynamically assess the reliability of pseudo-labels, thereby maximizing the utilization of unlabeled data.

Our main contributions can be summarized as follows:
\begin{itemize}
\item
To the best of our knowledge, the CTFS framework proposed in this paper is the first semi-supervised semantic segmentation framework specifically designed for the forward-looking sonar image. 

\item
We propose a collaborative teacher and multi-view reliability assessment algorithm for sonar image, which fully learn the characteristics of sonar images to maximally improve the student model’s semantic segmentation performance with extremely limited labels.

\item
We constructed a new Forward-Looking Sonar Semantic Segmentation (FSSG) dataset, which alleviates the scarcity of forward-looking sonar image datasets and their corresponding annotations.

\item
Extensive experiments conducted on the FLSMD and FSSG datasets demonstrate that CTFS consistently outperforms current state-of-the-art methods on mIoU.
\end{itemize}


\begin{figure*}[t]
    \centering
    \includegraphics[width=\linewidth]{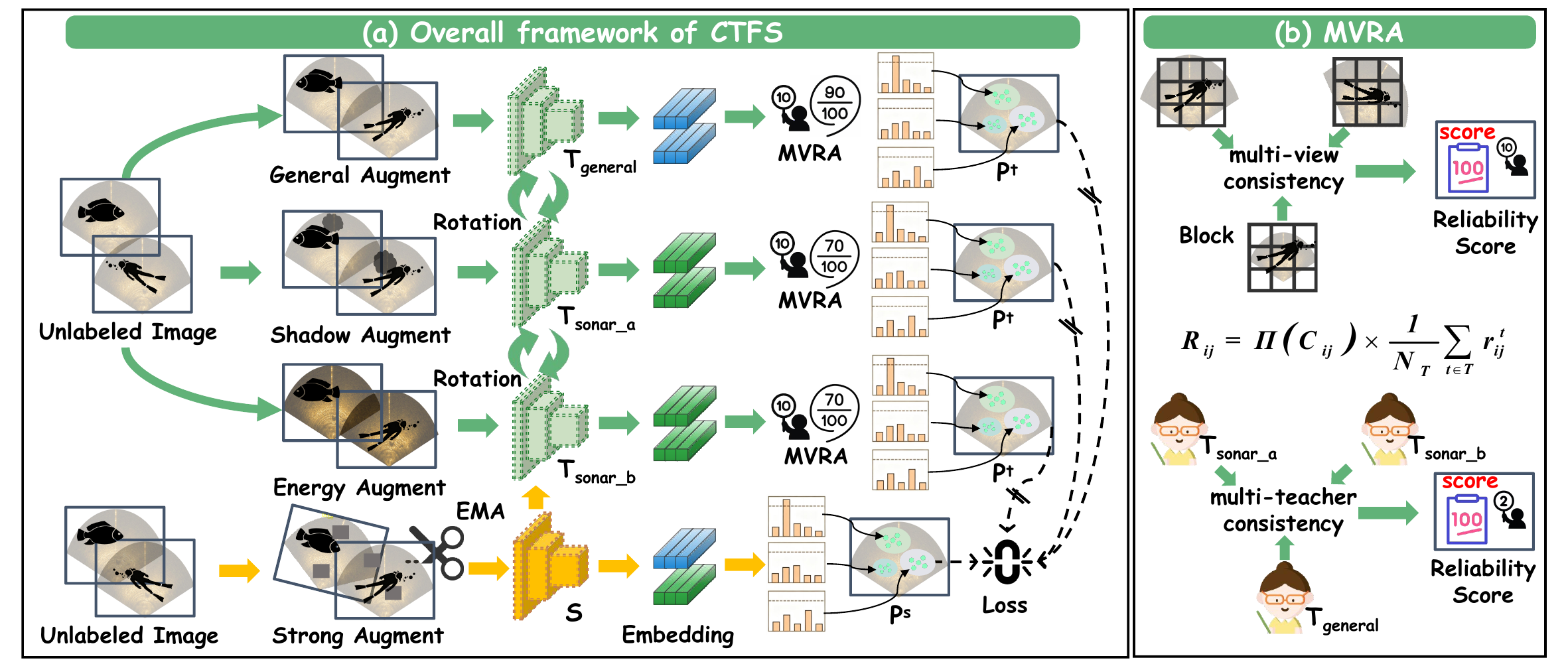}  
    \caption{(a) The overall architecture of CTFS, where knowledge is transferred to the student through the collaboration between the traditional teacher and the sonar teacher.
(b) The multi-view reliability assessment process of pseudo-labels.}

\label{fig:Sonar2}
\end{figure*}

\section{Related Work}
\label{sec:formatting}

\subsection{Semi-Supervised Semantic Segmentation}
The goal of semi-supervised semantic segmentation is to leverage a small number of labeled pixels to guide the learning of a large amount of unlabeled data, thereby obtaining stable and well-generalized decision boundaries~\cite{hu2021semi,hoyer2024semivl,lu2025improving,wang2025multi}. Mean Teacher uses exponential moving averages (EMA) to construct a smooth teacher to improve pseudo-label stability and robustness~\cite{tarvainen2017mean}. FixMatch unifies weak augmentation to generate high-confidence pseudo-labels and strong augmentation alignment learning into a concise threshold strategy.~\cite{sohn2020fixmatch}. After migrating to the pixel level, pseudo-label noise and boundary uncertainty become bottlenecks: PseudoSeg cleanses labels through confidence and boundary constraints~\cite{DBLP:conf/iclr/ZouZZLBHP21}. To alleviate bias and class imbalance, CPS employs dual-branch cross-pseudo-supervision~\cite{chen2021semi}, CCT improves robustness with diverse augmented prediction alignment~\cite{ouali2020semi}, and CutMix/ClassMix enhance pseudo-label stability and sample diversity under strong perturbations~\cite{yun2019cutmix,olsson2021classmix}. Meanwhile, strong representational encoders and self-supervised pre-training~\cite{dosovitskiy2020image,he2022masked,caron2021emerging,oquab2023dinov2} provide more separable features and richer context, working in conjunction with a concise and consistent Unimatch series of frameworks~\cite{yang2023revisiting,yang2025unimatch}. However, existing studies mostly validate on natural benchmarks, failing to adequately consider the strong speckle noise, acoustic shado, and salient domain inadequacy of sonar image.

\subsection{Benchmarks}
General semantic segmentation benchmarks such as PASCAL VOC 2012~\cite{everingham2015pascal}, Cityscapes~\cite{cordts2016cityscapes}, ADE20K~\cite{zhou2019semantic}, and COCO/COCO-Stuff~\cite{lin2014microsoft,caesar2018coco} have established standardized annotation and evaluation specifications. FLSMD~\cite{singh2021marine}, as one of the few publicly available corpora for the extremely scarce  forward-looking sonar image, which is consist of 12 target categories, covers scenes such as watertank, turntable, and flooded-quarry. It supports detection, classification, and semantic segmentation evaluation with unified data release and task definition. However, significant imaging differences exist among different sonar devices. The scarcity of existing sonar data and their corresponding semantic segmentation annotations severely limits the robustness and generalization capability of models. To further promote research on semantic segmentation in the sonar domain, we constructed and released a new dataset, FSSG, and provided it with paired semantic segmentation annotations.

\section{Methodology}
\label{sec:formatting}
\subsection{Overview}
Figure~\ref{fig:Sonar2} illustrates the overall framework of CTFS. This framework adopts a teacher-student model training strategy: in each training cycle, a certain number of labeled and unlabeled images are randomly selected as the current training data batch based on the data sampling ratio. Labeled data is input into the student model to calculate the supervised loss \( \mathcal{L}_{\text{sup} }\) with its ground truth, and unlabeled data is input into teacher model and student model respectively after weak enhancement and strong enhancement. The prediction results of the teacher model serve as pseudo-labels to supervise the student model, thereby calculating the unsupervised loss \( \mathcal{L}_{\text{unsup} }\). The training objective of the student model is as follows:
\begin{equation}
\mathcal{L}_{\rm total} = \mathcal{L}_{\rm sup} + \lambda_{u} \cdot \mathcal{L}_{\rm unsup},
\label{eq:total_loss}
\end{equation}
where \(\lambda_u\) is the hyper-parameter to adjust the contribution of the unsupervised loss. And the supervised loss \(\mathcal{L}_{\text{sup} }\) trained with a small amount of labeled data is as follows:
\begin{equation}
\mathcal{L}_{\rm sup} = \frac{1}{N_{l}} \sum_{n=1}^{N_{l}} \frac{1}{N_p}\sum_{b=1}^{H \times W} \mathcal{L}_{CE}\big(p_{b}^{n},y_{b}^{n}\big),
\label{eq:Ls_sup}
\end{equation}
where $N_l$ and $N_p$ denote to the number of labeled data and the number of whole pixels in every labeled data respectively. H and W refer to the pixel height and width respectively. \( p^{n}_{b} \) is the predicted category probability vector of the student model for the {\em b-th} pixel in the {\em n-th} image and \( y^{n}_{b} \) is the true label of the corresponding pixel. \(\mathcal{L}_{\rm CE}\) is the cross-entropy loss. The unsupervised loss \( \mathcal{L}_{\text{unsup} }\) will be discussed in the chapter on Reliability-Guided Adaptive Constraint.

\begin{figure*}[htbp]
    \centering
    \includegraphics[width=\textwidth]{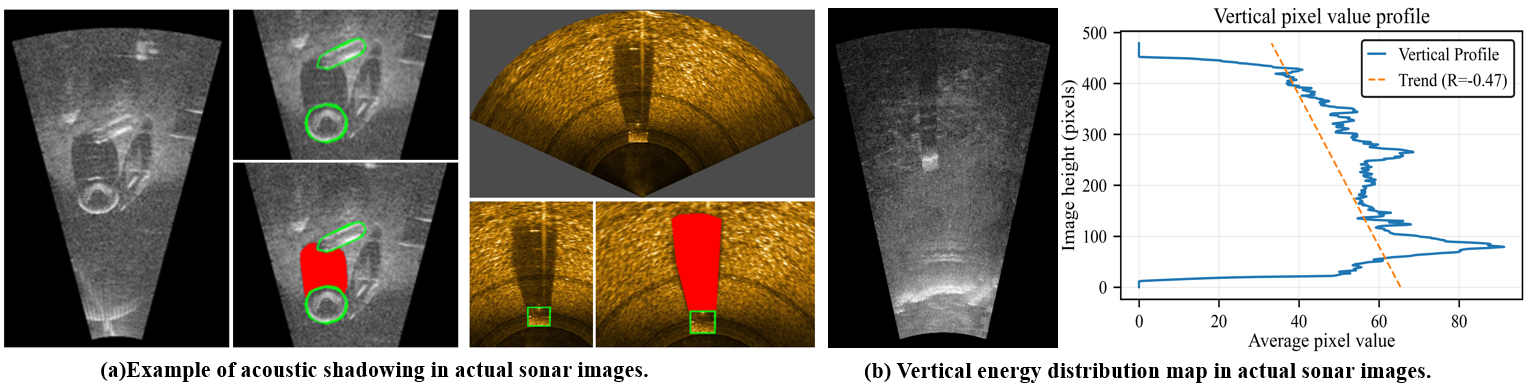} 
    \caption{(a) Example of shadows in sonar images: due to the obstruction of objects during sonar propagation, a shadow is formed behind the object. The green hollow area represents the GT, and the red area represents the shadow behind the target. (b) Example of how sonar energy decreases with propagation distance. In the right-hand image, the solid blue line represents the vertical energy distribution in the left-hand image, and the dashed red line is a linear regression fit to the blue line.}
    \label{fig:Simulate}
\end{figure*}

\subsection{Collaborative Teacher for Sonar(CBTS)}
Traditional teacher-student semi-supervised methods exploit the consistency between weak and strong enhancements to extract semantic information from unlabeled data. Weak enhancement operations used by teachers, such as color jittering for optical images, usually cannot effectively utilize the features of sonar image, and this knowledge gap accumulates with the EMA update mechanism, affecting the final model performance. To address this, we propose the CTFS method, which introduces multiple dedicated teacher models that are cyclically activated at different knowledge feature levels to provide multi-dimensional supervision and guidance to the student model. As shown in Figure 2(a), this framework consists of one student model and three structurally identical teacher models. In each training cycle, these three teacher models are activated in a fixed order of "general → sonar\_a → sonar\_b" to guide the same student model, as shown below:
\begin{equation}\label{eq:teacher_schedule}
\phi(e)=
\begin{cases}
\text{supervised}, & e < E,\\[4pt]
T_{general}, & e \ge E\ \wedge\ (e - E)\bmod 3 = 0,\\[4pt]
T_{sonar\_a}, & e \ge E\ \wedge\ (e - E)\bmod 3 = 1,\\[4pt]
T_{sonar\_b}, & e \ge E\ \wedge\ (e - E)\bmod 3 = 2,
\end{cases}
\end{equation}
where $\phi(e)$ and $e$ denote the teacher rotation function and current training epoch respectively. $E$ denotes the number of warm up stage. $T$ denotes the set of teacher models, which includes general teacher $T_{general}$, sonar teachers $T_{sonar\_a}$ and $T_{sonar\_b}$.

Similar to previous studies, we use strong enhancement as the input to the student model and weak enhancement as the input to the teacher models to ensure the stability of pseudo-label quality. The prediction results of the student model for the unlabeled image $P_{u}^{s}$ are as follows:
\begin{equation}
P_{u}^{s}=\mathcal{P}\left(A_{s^{\prime}}^{s}\left(X_{u}\right)\right)
\label{eq:Conventonpseudo_label},
\end{equation}
where $\mathcal{P}$ and $A_{s^{\prime}}^{s}$ are the prediction function and the set of strong enhancement perturbation of the student model respectively. $X_{u}$ indicates the unlabeled images.

However, unlike the traditional single teacher teacher-student framework, different teachers activate with different perturbation combinations: $T_{general}$ uses traditional geometric transformations; $T_{sonar\_a}$ focuses on simulating environmental interference factors such as acoustic shadows. $T_{sonar\_b}$ focuses on simulating acoustic imaging characteristics such as energy attenuation effects. Their prediction results for unlabeled images $p_{u}^{t}$ are as follows:
\begin{equation}
p_{u}^t = \sum_{t \in T} \mathcal{P}_{\phi(e)} \!\left( A_{w}^{t}\!\left( X_{u} \right) \right),
\label{eq:pseudo_label}
\end{equation}
where $t$ denotes the currently active teacher model and $A_{w}^{t}$ denotes the set of weak enhancement perturbation of the teacher model $t$ respectively.

Specifically, the acoustic shadow perturbation process we designed for $T_{sonar\_a}$ is represented as follows:
\begin{equation}
I_{o}(x,y)=
\begin{cases}
I_{i}(x,y)\times \bigl[1-\alpha\bigl(1-\frac{d}{R}\bigr)\bigr], & \text{if}\ \ (x,y)\in\mathcal{S}, \\
I_{i}(x,y), & \text{otherwise},
\end{cases}
\end{equation}
where $I_{o}(x,y)$ denotes the output of the pixel $(x,y)$ after applying shadow augment and $I_{i}(x,y)$ denotes the original pixel $(x,y)$. $\alpha$ is the shadow Intensity hyperparameter and $d$ is the distance between the current pixel and the starting point pixel of the shadow region. $\mathcal{S}$ and $R$ denote the shadow area and the maximum shadow radius respectively, and defined as follows:
\begin{equation}
\mathcal{S} = \left\{(x,y) \mid \theta \leq \phi(x,y) \leq \theta + \Delta\theta \ \land\ d(x,y) \leq R\right\},
\end{equation}
\begin{equation}
R = 0.2 \times \min(H, W),
\end{equation}
where $\theta$ and $\Delta\theta$ denote the starting angle and angular span of the shadow area respectively. $\phi(x,y)$ and $d(x,y)$ denote the angle and distance between the pixel $(x,y)$ and the shadow start pixel respectively, and defined as follows:

\begin{figure*}[t]
    \centering
    \includegraphics[width=\linewidth]{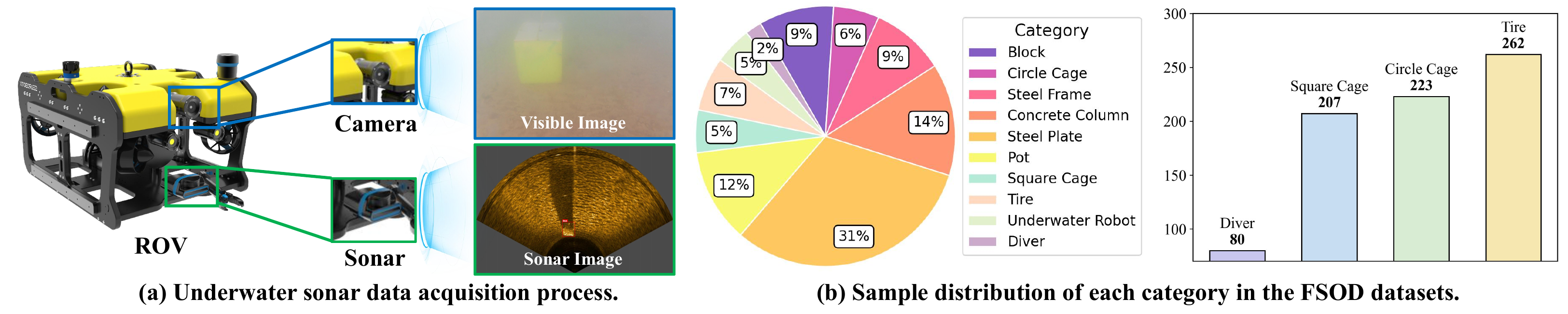}  
    \caption{(a) The collection process of the FSSG dataset. (b) Sample distribution of each category in the FSSG dataset, and the visualization of targets under sonar and visible perspectives.}

\label{fig:Sonar3}
\end{figure*}

\begin{equation}
\phi(x,y) = \arctan(y - y_0, x - x_0),
\end{equation}
\begin{equation}
d(x,y) = \sqrt{(x - x_0)^2 + (y - y_0)^2},
\end{equation}
where $(x_0,y_0)$ denotes the start point pixel of the shadow area. This design is intended to simulate the shadow formed when sonar is blocked by an obstruction during propagation, as shown in Figure~\ref{fig:Simulate}(a). And the energy attenuation perturbation process we designed for $T_{sonar\_b}$ is represented as follows:
\begin{equation}
I_{o}(x,y) = I_{i}(x,y) \times (1 - \gamma \times \frac{y}{H}),
\end{equation}
where $\gamma$ and $y$ denote the attenuation intensity hyperparameter and the vertical coordinate of the pixel respectively. This design aims to simulate the directional energy attenuation caused by factors such as seawater absorption of sound wave beams, as shown in Figure~\ref{fig:Simulate}(b).

Mapping each teacher to a specific perturbation preserves feature-level semantics, enabling the student to stably learn teacher-specific knowledge from strongly perturbed inputs and progressively transferring from general capability to domain specialization, thereby mitigating the poor cross-domain adaptability of single-teacher methods.
\subsection{Multi-view Reliability Assessment(MVRA)}

The conventional single confidence threshold method often fails to accurately identify high-quality pseudo-label. To address this, we propose a multi view reliability assessment mechanism(MVRA), as shown in Figure~\ref{fig:Sonar2}(b). This mechanism is independent of the pseudo-label generation process and consists of two core parts: single-teacher intrinsic stability verification and cross-teacher consistency verification. Specifically, we first segment the image into multiple grid blocks, each with same size of $m$, and then calculate the average of the predicted category probability vectors of all pixels in each grid to act as the category probability of that grid, as shown below:
 \begin{equation}\label{eq:fkij_m}
f^{t}_{ij} \;=\; \frac{1}{N_{\Omega_{ij}}} \sum_{(u,v)\in \Omega_{ij}} p^{t}(u,v),
\end{equation}
where $\Omega_{ij}$ and $N_{\Omega_{ij}}$ denote the set of pixels in the {\em (i,j)-th} grid and the total number of pixels in this set respectively. $p^{t}(u,v)$ is the category probability vector prediction of the teacher model $t$ for the pixel $(u,v)$.

 In single-teacher intrinsic stability verification, each teacher assesses the reliability of their prediction by testing the consistency between different augmented views of the same image and the original image. The detailed calculations are as follows:
 \begin{equation}\label{eq:r_ij_m}
r_{ij}^{t} \;=\; \frac{1}{N_{A_{w}^{t}}} \sum_{k=1}^{A_{w}^{t}} \cos\!\bigl(f_{ij}^{ot},\,f_{ij}^{kt}\bigr),
\end{equation}
where $r_{ij}^{t}$ indicates the prediction stability score of teacher $t$ for the {\em (i,j)-th} grid patch. $\cos$ denotes the cosine similarity. $f_{ij}^{ot}$ and $f_{ij}^{kt}$ are the category probability vector predictions of the teacher $t$ for the {\em (i,j)-th} grid patch in the original image and the augment view k respectively.
 
 In cross-teacher consistency verification, three teachers simultaneously generate predictions for the same unlabeled image, and the reliability of each grid patch is assessed based on the consistency among their outputs. This process is described as follows:
\begin{equation}
C_{ij} = \frac{1}{N_{\mathcal{D}}} \sum_{(p,q)\in\mathcal{D}} \cos\!\left( f_{ij}^{op},\, f_{ij}^{oq} \right),
\end{equation}
where $C_{ij}$ denotes the multi-teacher prediction consistency score for the {\em (i,j)-th} grid patch. $\mathcal{D}$ and $N_{\mathcal{D}}$ are the set of teacher pairs and the number of the pairs respectively. $p$ and $q$ represent different teacher models respectively.
 
 Finally, by fusing the intrinsic stability of each single teacher and the predictive consistency of multiple teachers, we obtain a reliability score for each grid cell and then replicate it to each pixel within that grid cell to generate a pixel-level reliability score, as shown in the following calculation:
 \begin{equation}\label{eq:Reliability}
R_{ij} \;=\; \Pi\big(C_{ij}\big)\;\times\frac{1}{N_T}\sum_{t\in T}r_{ij}^{t},
\end{equation}
 where $N_T$ denotes the number of teacher models and $\Pi\big(C_{ij}\big)$ denotes the Penalty item based on consistency of multi-teacher predictions, which is defined as follows:
 \begin{equation}\label{eq:pi_Cij}
\Pi\big(C_{ij}) = \delta + (1-\delta)\,C_{ij},
\end{equation}
where $\delta$ denotes the consistency smoothing hyper-parameter to adjust the severity of punishment based on the consistency of predictions among multiple teachers.


\begin{figure*}[t]
    \centering
    \includegraphics[width=\linewidth]{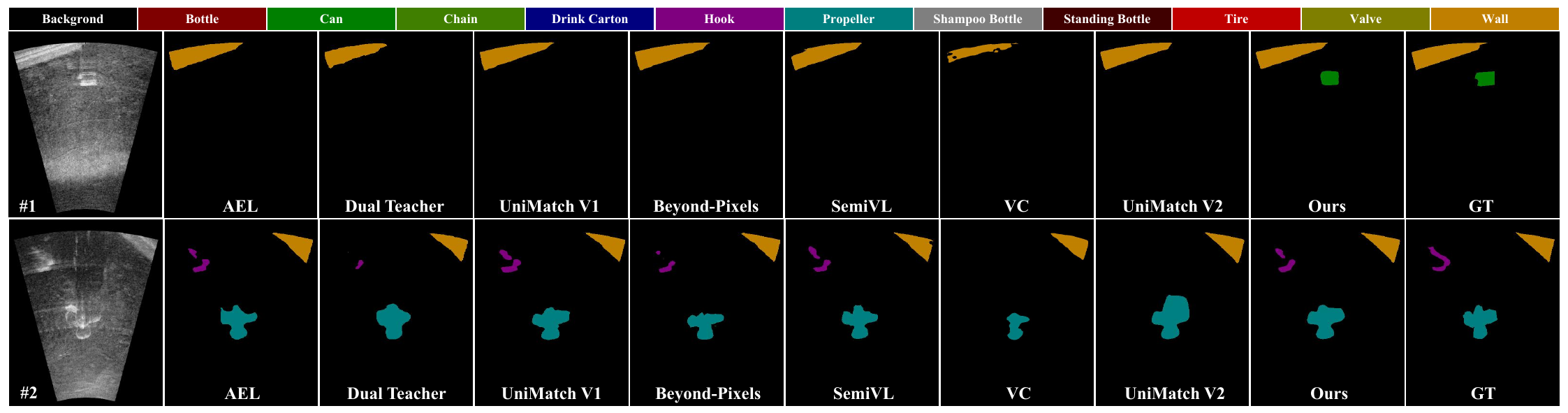}  
    \caption{Qualitative demonstrations of different approaches on the  FLSMD dataset with 2\% labeled data.}

\label{fig:Sonar10}
\end{figure*}

\begin{figure*}[t]
    \centering
    \includegraphics[width=\linewidth]{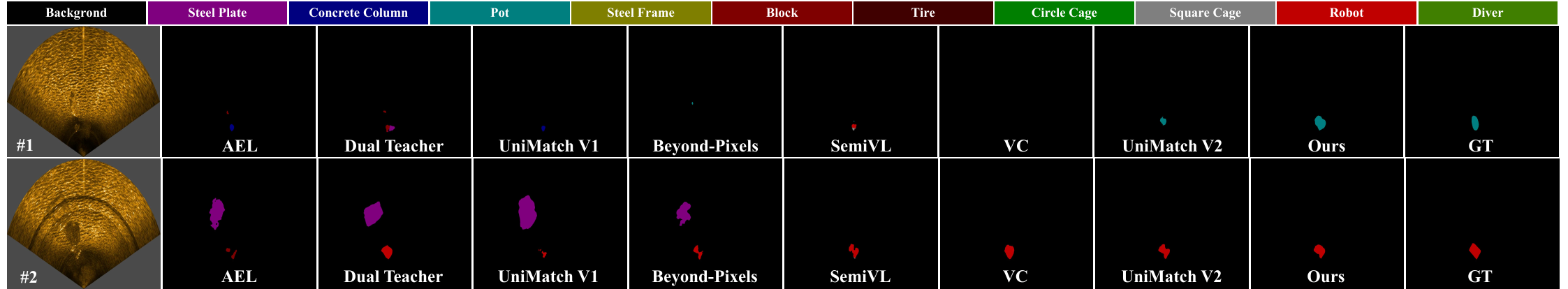}  
    \caption{Qualitative demonstrations of different approaches on the  FSSG dataset with 2\% labeled data.}

\label{fig:Sonar11}
\end{figure*}

\begin{table*}[t]   
  \centering
  \caption{Quantitative comparison of mIoU (\%) on the FLSMD and FSSG datasets.
  Experiments on both the FLSMD and FSSG datasets were conducted with labeled data of 2\%, 5\%, and 10\%. 
  The best results are shown in \textbf{bold}, and the second best results are shown in \underline{\textit{italic}}.} 
  \setlength{\tabcolsep}{10pt}
  \label{tab:comp1}
  \begin{tabular}{c|c|c|ccc|ccc}  
    \toprule 
    \multirow{2}{*}{\textbf{Method}} & 
    \multirow{2}{*}{\textbf{Venue}} & 
    \multirow{2}{*}{\textbf{Encoder}} & 
    \multicolumn{3}{c|}{\textbf{FLSMD}} & 
    \multicolumn{3}{c}{\textbf{FSSG}}\\
    & & & 2\% & 5\% & 10\% & 2\% & 5\% & 10\%\\
    \midrule
    Labeled Only & -- & RN-101    
          & 46.47 & 55.15 & 61.94 & 17.64 & 32.23 & 52.27\\
    Labeled Only & -- & DINOv2-S    
          & 51.08 & 61.02 & 68.48 & 35.61 & 41.67 & 54.58\\
    \midrule 
    AEL~\cite{hu2021semi} & NIPS’21 & RN-101  
          & 52.70 & 64.89 &  \underline{\textit{70.73}} & 53.51 & 57.84 & 63.08\\    
    Dual Teacher~\cite{na2023switching} & NIPS’23 & MiT-B2  
          & 56.23 & 66.09 & 69.61 & 58.31 & 60.16 & \underline{\textit{65.91}}\\  
    UniMatch V1~\cite{yang2023revisiting} & CVPR'23 & RN-101  
          & 50.70 & 56.14 & 61.66 & 58.39 & 60.87 & 63.15\\  
    Beyond-Pixels~\cite{howlader2024beyond} & ECCV'24 & RN-101  
          & 48.19 & 55.52 & 60.49 & 57.20 & \underline{\textit{62.37}} & 64.52\\  
    SemiVL~\cite{hoyer2024semivl} & ECCV'24 & RN-101  
          & 53.38 & 65.16 & 70.02 & \underline{\textit{58.85}} & 62.19 & 65.24\\
    VC~\cite{chen2024virtual} & TPAMI'24 & RN-101  
          & 50.53 & 59.36 & 66.86 & 49.76 & 54.75 & 59.27\\
    UniMatch V2~\cite{yang2025unimatch} & TPAMI'25 & DINOv2-S  
          &  \underline{\textit{57.24}} &  \underline{\textit{66.49}} & 69.81 & 58.78 & 61.21 & 64.41\\
    CTFS (Ours) & -- & DINOv2-S  
          & \textbf{62.32} & \textbf{68.08} & \textbf{72.27} 
          & \textbf{59.53} & \textbf{65.12} & \textbf{67.07}\\  
    \bottomrule  
  \end{tabular}
\end{table*}

\begin{figure*}[t]
    \centering
    \includegraphics[width=\linewidth]{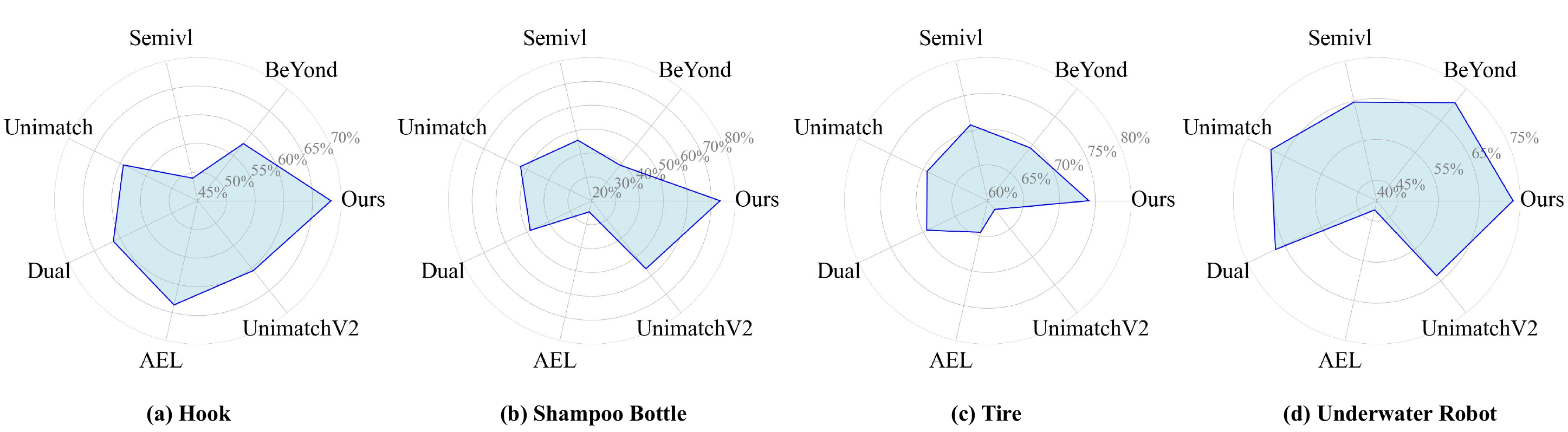}  
    \caption{Performance comparison of tail-class categories on the FLSMD dataset with a 2\% labeled and the FSSG dataset with a 5\% labeled. (a) performance comparison on the hook category, (b) performance comparison on the shampoo-bottle category, (c) performance comparison on the tire category, (d) performance comparison on the underwater robot category.}

\label{fig:Sonar7}
\end{figure*}

\subsection{Reliability-Guided Adaptive Constraint}
The multi-view reliability score evaluates prediction stability from multiple perspectives and is therefore more trustworthy than the conventional single-teacher prediction confidence. It can both replace the traditional prediction confidence—by applying a reliability threshold for hard filtering of high quality pseudo-labels, and be used as a soft weighting factor in the loss function to provide more fine-grained training guidance. Accordingly, the unsupervised loss in the CTFS framework is given by:
\begin{equation}
\mathcal{L}_{\mathrm{unsup}}
=\frac{1}{N_u}\sum_{n=1}^{N_u}
\frac{
\sum_{b=1}^{H\times W}
\mathcal{L}_{\mathrm{CE}}\!\left(p_{n}^{s}[b],\,p_{n}^{t}[b]\right)\,
\times R_{b}^{n}  \times \Delta\  
}{
N_p
},
\label{eq:unsup_loss}
\end{equation}
where $R_{b}^{n}$ denotes the reliability score of the {\em b-th} pixel in the {\em n-th} image. $\Delta$ is a binary vector used to determine whether a pixel participates in the loss calculation ,which can be expressed as follows:
\begin{equation}
\Delta = \begin{cases}
1 & \text{if } R_{b}^{n}  > {\psi},\\
0 & \text{otherwise}.
\end{cases}
\end{equation}
Where $\psi$ denotes the reliability score threshold.

 In this way, by introducing multiple teachers to address domain maladjustment issues, the CTFS framework’s multi-view reliability assessment also effectively mitigates the uneven pseudo-label quality caused by additional teachers, which conventional single-teacher confidence measures fail to filter accurately.
 
\section{Sonar semantic segmentation benchmark}
The scarcity of publicly available sonar datasets has severely constrained the development of sonar semantic segmentation. Collecting high-quality sonar semantic segmentation data is challenging due to high acquisition costs, reliance on specialized equipment, and complex underwater environments. To address these issues, we present the Forward-Looking Sonar Image Semantic Segmentation (FSSG) dataset, collected in Bohai Bay using a remotely operated vehicle (ROV) equipped with a multibeam forward-looking sonar (Oculus M750d), as shown in Figure~\ref{fig:Sonar3}(a). To reduce mutual interference and facilitate localization, objects were suspended by ropes at depths of approximately 3 to 20 meters and captured from different angles and ranges (2–15 meters) at 750 kHz and 1.2 MHz, increasing sample diversity.

After filtering, preprocessing, and annotation, the FSSG dataset contains 3761 images from 11 target categories. Figure~\ref{fig:Sonar3}(b) illustrates the distribution of minority categories. Notably, the steel plate category has the most samples (1229, 31\%), while the diver category has the fewest (80, 2\%), revealing the dataset’s inherent long-tailed distribution. Compared with the FLSMD dataset~\cite{singh2021marine}, FSSG additionally includes a diver category, providing valuable support for underwater rescue, accident prevention, and diving safety assurance systems.

\section{Experiment}
\subsection{Datasets and Experiment Settings}
\begin{table}[ht]   
\centering
\caption{Ablation study on key components of CTFS conducted on the FLSMD dataset with 2\% labeled data.}
\setlength{\tabcolsep}{5pt} 
\small
\resizebox{0.8\linewidth}{!}{
    \begin{tabular}{cccccc}  
        \toprule  
        Model & EMA & CBTS & MVRA & mIoU  \\
        \midrule  
        M1 & - & - & - & 51.08  \\
        M2 & $\surd$ & - & - & 55.94  \\
        M3 & $\surd$ & $\surd$ & - & \underline{59.94}  \\
        M4 & $\surd$ & $\surd$ & $\surd$ & {\textbf{62.32}}  \\
        \bottomrule  
    \end{tabular}
} 
\label{tab:comp3}  
\end{table}

\begin{table}[ht]
\centering
\small 
\caption{Ablation study on the effectiveness of the sonar-specific perturbations designed for the two additional sonar teachers on the FLSMD dataset with 2\% labeled data.}
\label{tab:pertur}
\begin{tabular}{c|c|c|c|c}
\toprule
\# & Method & mIoU & Hook & Shampoo Bottle \\
\midrule
1 & $T_{general}$ & 55.94 & 60.94 & 55.07 \\
2 & \#1 + $T_{sonar\_a}$ & 57.12 & 62.83 & 70.56 \\
3 & \#2 + $T_{sonar\_b}$ & 59.94 & 65.59 & 74.40 \\
\bottomrule
\end{tabular}
\end{table}

This study conducted comparative and ablation experiments on the FSSG dataset we collected and the publicly available FLSMD dataset~\cite{singh2021marine} which is widely used for sonar semantic segmentation~\cite{li2024lightweight}. Both datasets were split into training, validation, and test sets in a 6:2:2 ratio. Since there is currently no research on semi-supervised semantic segmentation for forward-looking sonar images, we followed the experimental protocols of related semi-supervised works~\cite{wang2023consistent}: 2\%, 5\%, and 10\% of the images from the training set were randomly sampled as labeled data while the remainder were treated as unlabeled data. In all experiments, we use the mean of Intersection over Union (mIoU) as a metric to evaluate the performance of semantic segmentation.

We use the simple DPT~\cite{ranftl2021vision} as our semantic segmentation model, which is built on DINOv2~\cite{oquab2023dinov2}. Specifically, weak augmentations for the conventional teacher include random resizing, random cropping, and horizontal flipping with a probability of 0.5~\cite{li2022pseco}. For the student model, strong augmentations include color jittering, grayscaling, and Gaussian blurring~\cite{tang2021humble}. During training, we employ the AdamW optimizer with a weight decay of 0.01. The learning rate for the pre-trained encoder is set to 5e-6, while that for the randomly initialized decoder is set to 2e-4. The entire model is trained on a single NVIDIA RTX 4090 GPU.

\subsection{Main Results}
In this section, we also compare it with several state-of-the-art semi-supervised semantic segmentation methods, including AEL~\cite{hu2021semi}, Dual Teacher~\cite{na2023switching}, UniMatch V1~\cite{yang2023revisiting}, Beyond-Pixels~\cite{howlader2024beyond}, SemiVL~\cite{hoyer2024semivl}, VC~\cite{chen2024virtual}, and UniMatch V2~\cite{chen2024virtual}. Using publicly available source code, we have reimplemented these benchmark methods on the proposed FSSG and FLSMD datasets.

\subsubsection{Qualitative Comparison}

As shown in Figure~\ref{fig:Sonar10}, all methods fail to detect the can in the first column, whereas our method successfully and completely identifies this target, maintaining the highest consistency with the ground truth—even when only 2\% of the data is labeled. This demonstrates that the proposed collaborative teacher architecture effectively captures the characteristics of sonar images and provides better support for the segmentation task, achieving results that other methods cannot. As shown in Figure~\ref{fig:Sonar11}, on the FSSG dataset, when only 2\% of the data is labeled, the segmentation results in the first column indicate that methods such as Dual Teacher and SemiVL produce incorrect segmentation, while VC fails to detect the target. In contrast, our method still achieves the best detection results, fully demonstrating its excellent generalization capability.

\subsubsection{Quantitative Comparison}

As shown in Table~\ref{tab:comp1}, CTFS significantly outperforms other methods on the FLSMD and FSSG datasets, demonstrating the effectiveness of our proposed collaborative-teacher and multi-view reliability assessmen algorithm. By introducing additional sonar teachers, the method transfers both unique and common feature knowledge to the student while effectively suppresses the influence of noisy pseudo-labels, thereby successfully enhancing the prediction accuracy of the student model. Notably, on the FLSMD dataset, our method achieves a 5.08\% improvement even with only 2\% of extremely scarce labeled data.

As shown in Figure~\ref{fig:Sonar7}, CTFS achieves significant performance advantages in terms of the mIoU metric for the tail classes hook and shampoo-bottle in the FLSMD dataset with a 2\% labeled data, as well as for tire and underwater robot in the FSSG dataset with a 5\% labeled data.

\subsection{Ablation Study}
\subsubsection{Component Analysis}

The results in Table~\ref{tab:comp3} validate the effectiveness of the key components in CTFS. M1 represents the performance of the DINOv2-S detector under fully supervised training, while M2 shows the results after introducing the teacher-student network. M3 further incorporates CBTS, and M4 presents the complete CTFS model with MVRA. The results demonstrate that both CBTS and MVRA contribute significantly to performance improvement. Through collaborative knowledge transfer between the professional sonar teacher and the traditional teacher, the student model is able to better capture the distinctive characteristics of sonar images, leading to substantial overall gains. Meanwhile, the proposed reliability evaluation strategy effectively alleviates the negative impact of noisy pseudo-labels caused by the unique properties of sonar images and the introduction of additional teachers.

\begin{figure}[t]
    \centering
    \includegraphics[width=\linewidth]{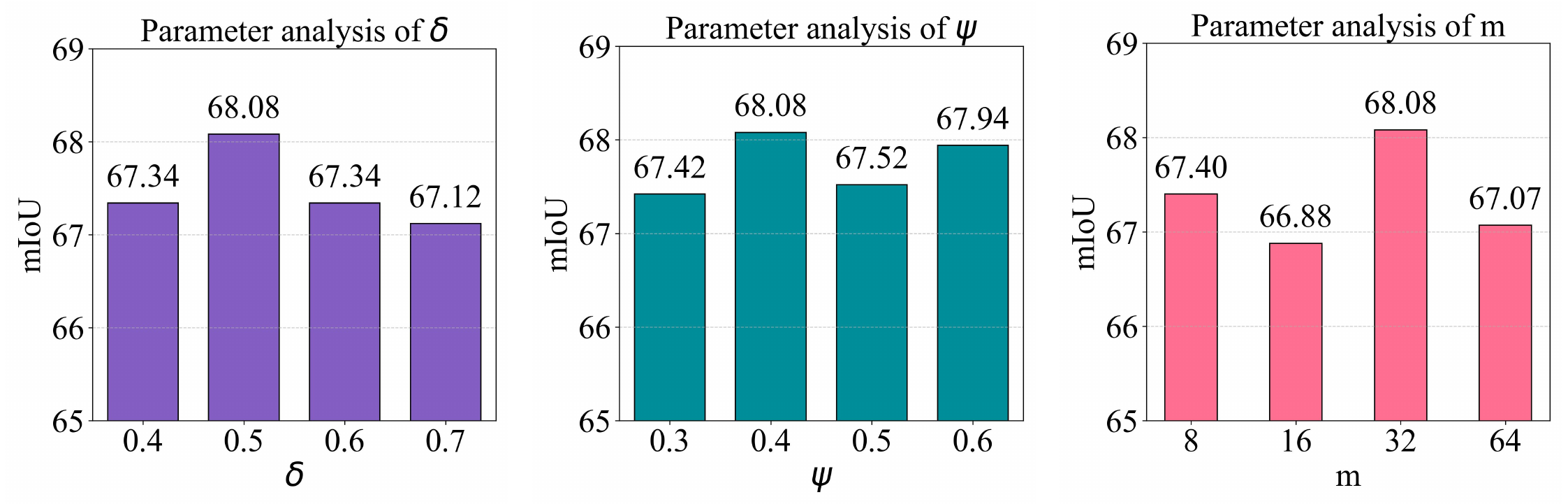}  
    \caption{mIoU results for each parameter on the FLSMD dataset with 5\% labeled data.} 
\label{fig:Sonar8}
\end{figure}

As shown in Table~\ref{tab:pertur}, we further analyze the effectiveness of the feature perturbations designed for the two additional sonar teachers, $T_{sonar_a}$ and $T_{sonar_b}$. Hook and Shampoo Bottle denote the two tail classes in the FLSMD dataset, accounting for only 4.4\% and 2.8\% of the total class instances, respectively. The results show that both sonar-specific perturbations bring significant and stable improvements across different aspects. This is because they take into account common real-world sonar imaging conditions, such as shadows behind objects caused by sound wave obstruction and energy attenuation resulting from seawater absorption and round-trip propagation.

\subsubsection{Parameter Analysis}
As shown in Figure~\ref{fig:Sonar8}, we analyzed the key parameters during the experimental process, including the consistency smoothing factor $\delta$, the reliability score threshold $\psi$ and the grid patch size $m$, noting that m=8 indicates an 8×8 grid. The experimental results show that the model achieves the best performance when the consistency smoothing factor $\delta$ is set to 0.5 and the reliability score threshold $\psi$ is set to 0.4, indicating that  $\delta$ of 0.5 can map the reliability score into a reasonable range which can effectively distinguish between high and low quality pseudo-labels. On this basis, $\psi$ of 0.4 effectively filters out low quality pseudo-labels, ensuring that the student model can continuously learn effective features from high quality pseudo-labels. Furthermore, experiments also show a 32×32 grid outperforms other sizes: smaller patches raise within-patch prediction variance, while larger patches include too many background pixels that falsely inflate predictive stability—both situations degrade the final performance.

\section{Conclusion}
We propose a solution for sonar image semantic segmentation with extremely limited labels. The framework introduces a collaborative teacher mechanism, allowing the teacher models to fully focus on the distinctive characteristics of sonar images and transfers knowledge to the student model by using unlabeled data. Guided by reliability scores, the student model can more accurately distinguish and learn from different pseudo-labels, significantly reducing the negative impact of noisy pseudo-labels during training. Extensive experiments conducted on two datasets demonstrate that CTFS exhibits remarkable superiority over other methods with extremely limited labels.

{
    \small
    \bibliographystyle{ieeenat_fullname}
    \bibliography{main}
}


\end{document}